\documentclass[letterpaper, 10 pt, conference]{ieeeconf} 
\IEEEoverridecommandlockouts      
\usepackage{times}
\usepackage[pdftex]{graphicx}
\usepackage{graphics}
\usepackage[cmex10]{amsmath}
\usepackage{url}
\usepackage{array}
\usepackage{tabularx}
\usepackage{multirow}
\usepackage{multicol}
\usepackage{booktabs}
\usepackage{epstopdf}
\usepackage{rotating}
\usepackage{csquotes}

\usepackage{amssymb}
\usepackage{amsmath}
\usepackage[all]{xy}    
\usepackage{ifthen}
\usepackage[usenames,dvipsnames]{color}
\usepackage{enumerate}
\usepackage{bm}
\usepackage{siunitx}
\usepackage{hyperref}

\usepackage{xargs}                      
\usepackage[pdftex,dvipsnames]{xcolor}  
\usepackage[colorinlistoftodos,prependcaption,textsize=tiny]{todonotes}
\newcommandx{\change}[2][1=]{\todo[linecolor=blue,backgroundcolor=white!25,bordercolor=blue,fancyline,#1]{#2}}
\newcommandx{\unsure}[2][1=]{\todo[linecolor=red,backgroundcolor=red!25,bordercolor=red,#1]{#2}}
\newcommandx{\info}[2][1=]{\todo[linecolor=OliveGreen,backgroundcolor=OliveGreen!25,bordercolor=OliveGreen,#1]{#2}}
\newcommandx{\improvement}[2][1=]{\todo[linecolor=Plum,backgroundcolor=Plum!25,bordercolor=Plum,#1]{#2}}
\newcommandx{\thiswillnotshow}[2][1=]{\todo[disable,#1]{#2}}

\makeatletter
\let\NAT@parse\undefined
\makeatother
\usepackage[numbers]{natbib}

\newcommand{\secref}[1]{Section~\ref{#1}}

\newcommand{\figref}[1]{Figure~\ref{#1}}

\newcommand{\myparagraph}[1]{\vspace{0.1in}\noindent\textbf{#1}}

\newboolean{draft-mode}
\setboolean{draft-mode}{true}
\newcommand{\sidenote}[1]{\ifthenelse{\boolean{draft-mode}}{\marginpar{\tiny\raggedright\textsf{\hspace{0pt}#1}}}{}}
\DeclareRobustCommand{\pynote}[1]{\ifthenelse{\boolean{draft-mode}}{\textcolor{green}{\textbf{PY: #1}}}{}}
\DeclareRobustCommand{\arnote}[1]{\ifthenelse{\boolean{draft-mode}}{\textcolor{blue}{\textbf{AR: #1}}}{}}
\DeclareRobustCommand{\nfnote}[1]{\ifthenelse{\boolean{draft-mode}}{\textcolor{red}{\textbf{NF: #1}}}{}}
\DeclareRobustCommand{\mbnote}[1]{\ifthenelse{\boolean{draft-mode}}{\textcolor{cyan}{\textbf{MB: #1}}}{}}

\title{\LARGE \bf Omnipush: accurate, diverse, real-world dataset \\ of pushing dynamics with RGB-D video
}
\author{
\authorblockN{Maria Bauza$^1$, Ferran Alet$^2$, Yen-Chen Lin$^2$, \\ Tom\'{a}s Lozano-P\'{e}rez$^2$, Leslie P. Kaelbling$^2$, Phillip Isola$^2$, and Alberto Rodriguez$^1$\\} \authorblockA{$^1$
    Mechanical
    Engineering Department --- Massachusetts Institute of Technology\\
    $^2$ Computer Science and Artificial Intelligence Laboratory ---
    Massachusetts Institute of Technology\\
    {\tt\small
      <bauza,alet,yenchenl,tlp,lpk,phillipi,albertor>@mit.edu}}\\
      \url{https://web.mit.edu/mcube/omnipush-dataset/}\\
    \includegraphics[width=\textwidth]{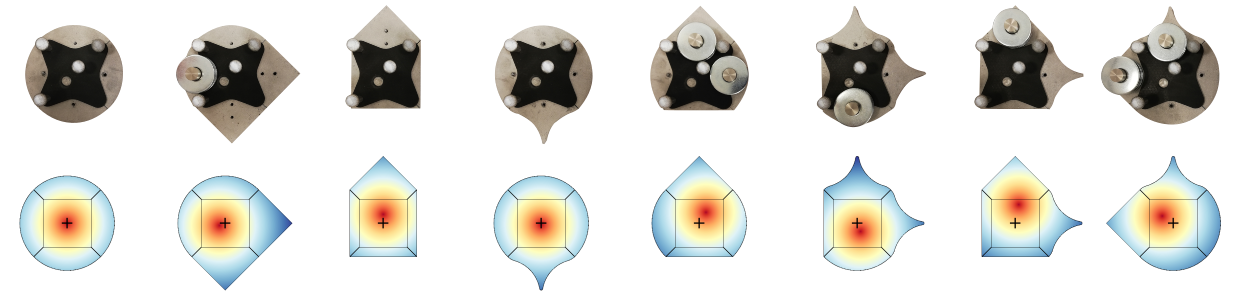} 
    \thanks{We want to thank Elliott Donlon for his help with the design and building the Omnipush objects; Angels Villalonga Riudavets for her assistance at generating the CAD model files, and Nick Walsh for helping to make this dataset publicly available.}
}

\begin{document}

\maketitle



\begin{abstract}
    Pushing is a fundamental robotic skill. Existing work has shown how to exploit models of pushing to achieve a variety of tasks, including grasping under uncertainty, in-hand manipulation and clearing clutter. Such models, however, are approximate, which limits their applicability.
    
    Learning-based methods can reason directly from raw sensory data with accuracy, and have the potential to generalize to a wider diversity of scenarios. 
    However, developing and testing such methods requires rich-enough datasets. In this paper we introduce Omnipush, a dataset with high variety of planar pushing behavior.
    
    In particular, we provide 250 pushes for each of 250 objects, all recorded with RGB-D and a high precision tracking system. The objects
    are constructed so as to systematically explore key factors that affect pushing --the shape of the object and its mass distribution-- which have not been broadly explored in previous datasets, and allow to study generalization in model learning.
    
    Omnipush includes a benchmark for meta-learning dynamic models, which requires algorithms that make good predictions and estimate their own uncertainty. We also provide an RGB video prediction benchmark and propose other relevant tasks that can be suited with this dataset.
     Data and code are available at \url{https://web.mit.edu/mcube/omnipush-dataset/}.
\end{abstract}

\section{Introduction}
\label{sec:introduction}

Object manipulation is central to robotics, but remains one of its most significant challenges. Among the possible ways to manipulate an object, pushing stands out as one of the most fundamental. On the one hand, pushing enables complex manipulation: reorienting objects, uncluttering scenes, and deforming objects. On the other hand, its simplicity makes pushing a good setting in which to explore technical advancements in robotic manipulation.

In earlier work, to improve our understanding of pushing and facilitate technical exploration, we developed a high-fidelity dataset of planar pushing experiments~\citep{yu2016more}.
Since its release, the earlier dataset has facilitated research in multiple research problems, which we review in~\secref{sec:relatedWork}. 
More importantly, feedback from users of the dataset has indicated that it could be improved by (a) providing realistic raw RGB-D sensor data in addition to tracking data, (b) adding increased diversity, and (c) creating a benchmark to evaluate generalization.

To address these limitations, we introduce the Omnipush\footnote{The name is inspired by the Omniglot dataset~\citep{Lake1332}. The Omniglot dataset
diversified the popular MNIST dataset, going from thousands of images for each of 10 characters to 20 images for each of 1623 characters.} dataset, with:
\begin{itemize}
\item Increased diversity, with 250 pushes for each of 250 objects. The previous dataset had thousands of pushes for each of 11 objects which was not well suited to study generalization across objects.
%
%
\item Controlled variation of the object's mass distribution. In the previous dataset, all objects had uniform mass distribution, leading to more homogenous dynamics.
\item State recorded both with RGB-D video as well as ground truth state tracking. The previous dataset only had ground truth state tracking.
\end{itemize}

Omnipush enables new research studies not supported by earlier datasets. The larger variety of shapes and mass distributions enables studying generalization; the difficulty of observing mass distribution enables tests of adaptive control, and RGB-D data enables studies on image and video prediction. Finally, the combination of intrinsic noise in robotic data with the epistemic uncertainty of small data domains enables research into meta-learning algorithms that model their own uncertainty. 

The paper is organized as follows: in \secref{sec:dataSet} we describe the main aspects of the dataset and in \secref{sec:influence} we illustrate the effect of shape and mass distribution on the dynamics of pushed objects. We also provide baselines on two possible applications: dynamic modeling in~\secref{sec:modeling} and video prediction  in \secref{sec:prediction};  and conclude in \secref{sec:conclusion} by discussing further potential applications and future extensions of the dataset.

\begin{figure}
\vspace{4pt}
  \begin{center}
    \includegraphics[width=\linewidth]{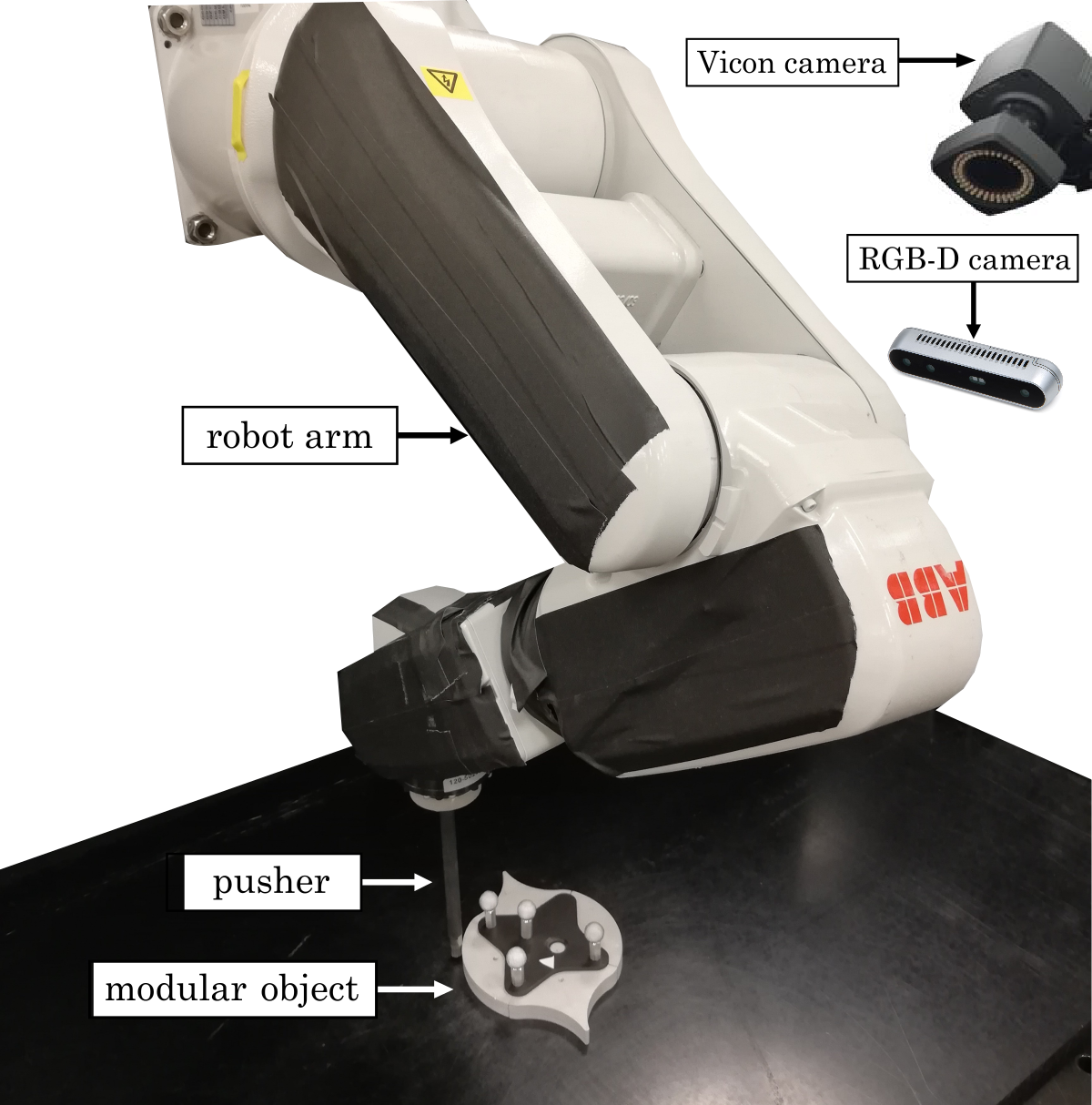}
  \end{center}
  \caption{\textbf{Data capturing setup.} The design of the system makes it possible to autonomously collect large amounts of data for each object. Human intervention is only required to assemble and replace objects.}
  \label{fig:hardware_robot}
\end{figure}

\section{Related Work}
\label{sec:relatedWork}

Robotics work on planar pushing goes back to the late 80s and early 90s~\citep{mason1986mechanics, Goyal1991,lee1991fixture} where model-based developments such as the voting theorem~\citep{mason1986mechanics} and the limit surface~\citep{Goyal1991} established the foundation of its mechanics.
Since then, several model-based methods have been proposed to describe the dynamics of pushed objects~\citep{lynch1992manipulation,howe1996practical,lynch1996stable,peshkin1988motion}. At the same time, recent work has shown that the assumptions used in these models fail to hold in a variety of real scenarios \citep{yu2016more,bauza2017probabilistic}.

On the other hand, there has been extensive work on learning pushing models from data~\citep{Salganicoff1993,Walker2008,Lau2011,Meric2015,Zhou2016}, including characterizing the stochasticity of planar pushing~\citep{bauza2017probabilistic}, improving physics-based models with data~\citep{kloss2017combining,ajayIROS2018} and learning dynamics from raw inputs~\citep{finn2017deep,agrawal2016learning,kloss2017combining}. Others have demonstrated the potential use of learned dynamic models for planning and control~\citep{zeng2018learning,finn2017deep,peng2018sim,clavera2017policy}.

Two of these projects~\citep{finn2017deep,agrawal2016learning} provide RGB datasets of pushing. However, they do not provide depth information and lack accurate pusher and object pose tracking. They provide data for a wider variety of objects, but the variability is not systematic, making it difficult to study. 

Since we presented our earlier dataset on planar pushing~\citep{yu2016more}, it has been directly used for: 

\begin{enumerate}
    \item Stochastic modeling: \cite{zhou2018convex,bauza2017probabilistic, ajayIROS2018}
    \item Modeling from rendered images: \cite{kloss2017combining}
    \item Model identification: \cite{zhu2018}
    \item Learning models for control: \cite{baumeistercombining,bauza2018}
    \item Filtering: \cite{bauza2017gp}
    \item Meta-learning: \cite{alet2018modular}
\end{enumerate}

With this new dataset we hope to further facilitate research in learning models and control.


\section{The Omnipush Dataset}
\label{sec:dataSet}

In this section we describe the main properties of the Omnipush dataset. First, we detail the data-collection setup, including the robot, the pusher, the planar surface, the high-fidelity tracking system, the RGB-D camera and the software used to record the data. Next, we introduce the set of pushed objects and their main properties. We also provide notation to uniquely refer to each object and explain the criteria used to decide their shape and mass distribution. Finally, we describe the process of collecting pushes and extensions to the dataset.  

\subsection{Data collection system}\label{sec:system}


The pushing system used for the data collection, shown in~\figref{fig:hardware_robot}, is based on an industrial robot arm that pushes a given object over a flat surface. We introduce variations on the pushed objects to study how the dynamics of pushing change depending on the object and keep the rest of the setup constant during the experiments.
The main parts of the system are:

\myparagraph{Robot and pusher.}
The system uses an ABB IRB 120 industrial robotic arm with 6 DOF to precisely control the position and velocity of the pusher. The pusher is a stiff steel rod moved perpendicular to the surface. The pusher has a length of 156~mm and a diameter of 9.5~mm, which minimizes occlusions while providing enough rigidity. The pose of the pusher is directly given by the robot, with an estimated accuracy of 0.1~mm.

\myparagraph{Surface.}
The surface where the frictional interaction occurs is made of ABS, a hard plastic with coefficient of friction of around 0.15. We selected this surface as it provides consistent friction both spatially and over time, due to its resistance to wear.\looseness=-1

\myparagraph{Motion tracker.}
We track the pose of the object with a Vicon motion tracking system, composed of 4 Bonita cameras. Each object carries 4 reflective markers that cameras detect to estimate the object pose. This system is very accurate and provides object pose estimations with an accuracy that can reach 0.5~mm for translation and 0.5$^\circ$  for rotation.  \looseness=-1

\myparagraph{RGB-D camera.}
RGB-D images are recorded using an Intel Realsense Camera D415 rigidly mounted and looking towards the workspace of the robot. 
RGB and depth are aligned and recorded at a frequency of 30Hz and at a resolution of 640x480.\looseness=-1

\myparagraph{Software.}
We integrate the components of the system, such as robot control, motion tracking and RGB-D recording, using the Robot Operating System (ROS). The data streams of robot pose and object pose from the Vicon system are published as ROS topics and recorded at 250~Hz while RGB-D images are published at 30~Hz.  The experiments are logged to ROS bag files, and we also provide them in HDF5 and JSON formats. Refer to \url{https://web.mit.edu/mcube/omnipush-dataset/} for more details and code. \looseness=-1

\subsection{Omnipush shapes}\label{sec:shapes}

\begin{figure}
\vspace{4pt}
\centering
\includegraphics[width=\linewidth]{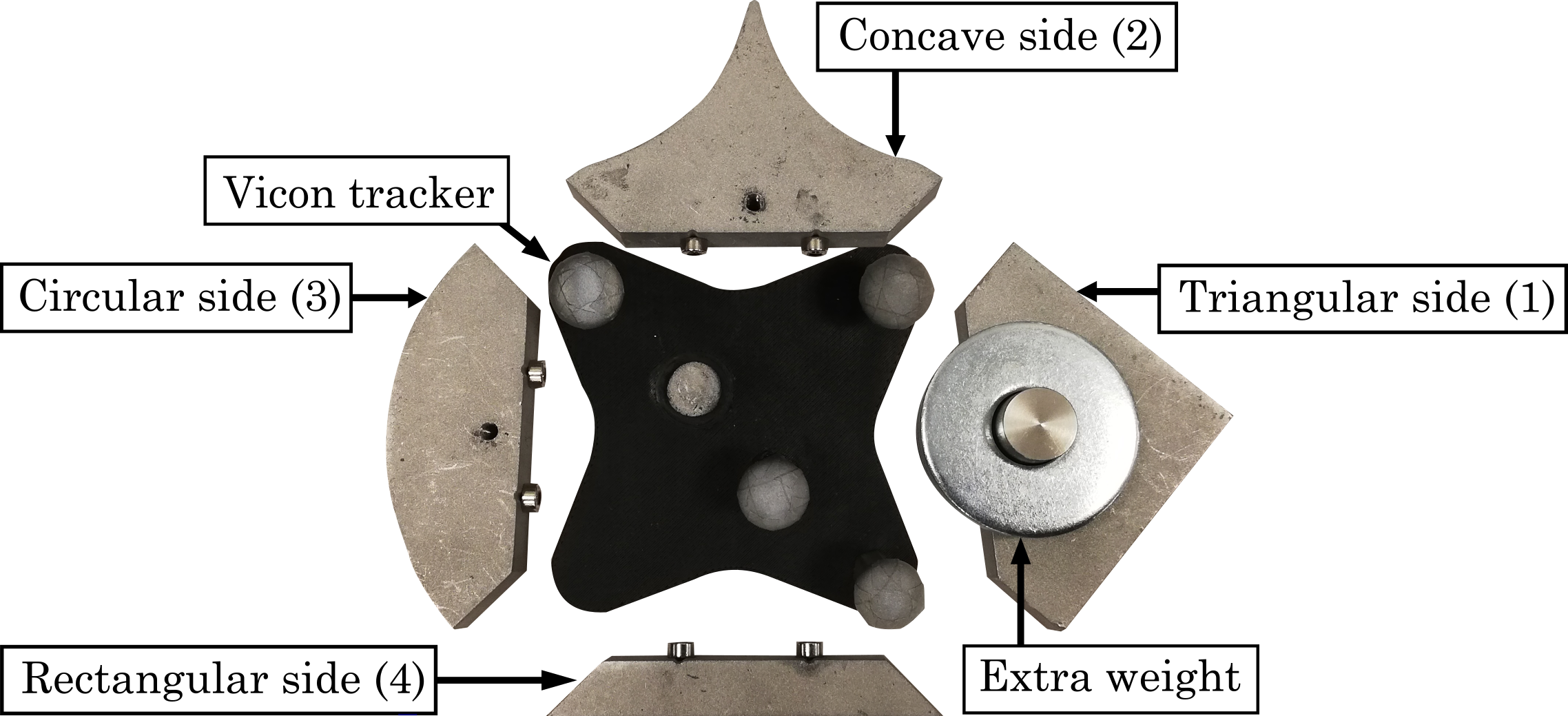}
\caption{\textbf{Modular object.} Object made of the four possible sides and named '1a3a4a2b' following the dataset convention. Weights can be added to the holes. There is an extra hole in the triangular side, covered by the wafers.}
\label{fig:sample_object}
\vspace{-5mm}
\end{figure}

The Omnipush objects are built modularly from a set of 6061 aluminium pieces as shown in~\figref{fig:sample_object}. All objects share a square central piece (100.5g, made of PLA) that carries the Vicon markers. Since they are placed on the shared piece, all puzzle-built objects are tracked with the same accuracy. To this central piece, we magnetically attach 4 different sides. These sides are locked in the horizontal plane and have freedom to move in the vertical direction so that they lay flat on the support surface and contribute to the frictional interactions. Each side is selected from a set of four different types, leading to 70 different shapes with nearly uniform mass distribution (Vicon markers and magnets have almost negligible weight). The four possible sides are concave (74.1g), triangular (94.1g), circular (67.5g) and rectangular (31.2g); which we denote with numbers \texttt{1-4} respectively. \figref{fig:sample_object} shows an example object made with each possible side.


To add more diversity, we change the center of friction of some objects by altering their mass distribution. All sides except the rectangular one allow the addition of extra weight at 35~mm from the center of the square piece. We considered two different extra weights: small (\texttt{b}, 60g) and large (\texttt{c}, 150g). No extra weight is denoted as \texttt{a}. The triangular side allows the extra weight to be placed in two different positions (interior, denoted as \texttt{b} or \texttt{c}, and exterior weight, at 50~mm, denoted as \texttt{B} or \texttt{C}). The biggest weight is around 5 times heavier than the smallest side.

The set of 250 objects consists of three groups, depending on the extra weights added:
\begin{itemize}
    \item \textbf{70 objects without an extra weight.}  We included \textit{all} objects, up to rotational symmetries, that can be assembled using four different sides and no extra weights.
    \item \textbf{90 objects with one extra weight.} We collected 45 objects with a small weight and the same 45 shapes with a large weight. For example, if we randomly selected the object \texttt{1a3a4a2b} shown in~\figref{fig:sample_object}, then we also included the object \texttt{1a3a4a2c}. 
    \item \textbf{90 objects with two small extra weights.} We randomly select objects that contain two small weights. We get objects of the form: \texttt{1b1b3a2a} and \texttt{1a2B2a3b}.
\end{itemize}

The CAD models of all 250 objects are available online.

\subsection{Data collection process}\label{sec:collection}

The data collection is autonomous and independent of the object. Given an object to explore, we collect data of its pushing dynamics by following this scheme:

\begin{enumerate}
    \item Move the pusher at a random position between 9 and 10~cm from the center of the object's central piece. This prevents the pusher from starting in a position that collides with the object, regardless of its shape.
    \item Select a random direction and make a 5cm straight push for 1s. The velocity is constant and chosen so that the interaction is close to quasistatic, meaning that the object stops moving as soon as the robot stops pushing it \citep{bauza2017probabilistic}. Repeat 5 times without changing the pusher position between the end of one push and the start of the next.\looseness=-1
    \item Go back to step 1.
\end{enumerate}

This scheme applies to all pushes unless, during step 2, the object ends outside a predefined region of the workspace. In that case, we stop pushing the object and the robot pulls it back to the center of the workspace and data collection restarts at step 1. 

To ensure a random distribution of pushes, the data collection setup is not aware of the shape of the object, which leads to a sizable portion of pushes resulting in no contact. To increase the frequency of pushes with contact, we modify the strategy for randomly selecting the direction of each push. If there was contact in the previous push and the pusher starts from where the previous push ended, then with probability 0.8 we select an angle that deviates at most $\pm90^\circ$ from the previous pushing direction. Otherwise, the direction of pushing is uniformly sampled across all angles, filtering out those that end more than 15~cm from the center of the object.\looseness=-1

In total, we collected 250 pushes for each of the 250 objects, making more than 60k accurately recorded pushes. 
We believe this dataset contains a diverse and complex set of examples that are sufficient to capture some of the most fundamental characteristics of pushing, such as the effect of different pressure distributions and object shapes. 
Each push recorded contains the following:

\begin{figure}\centering
\vspace{4pt}
\includegraphics[width=3in]{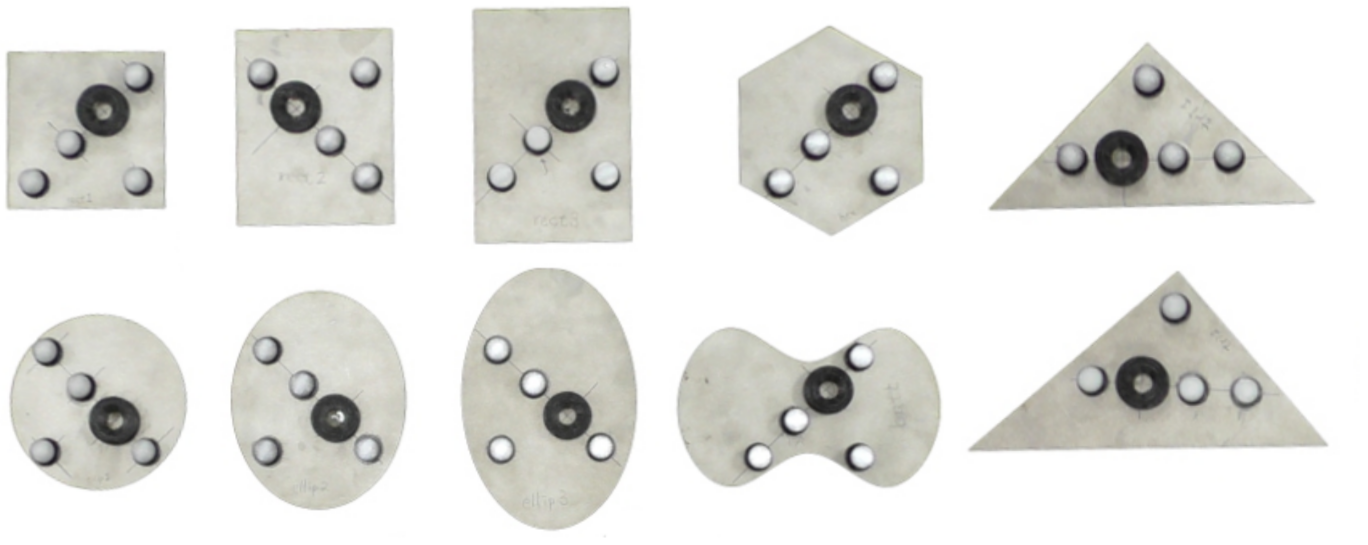}
\caption{\textbf{Out-of-distribution objects.} 10 objects from the previous dataset~\cite{yu2016more}, made of a different material (stainless steel), are much heavier and have different shapes compared to the 250 modular objects.}
  \label{fig:out_objects}
  \vspace{-1\baselineskip}
\end{figure}

\begin{figure*}[ht]
\vspace{4pt}
\centering
{\includegraphics[width=\linewidth]{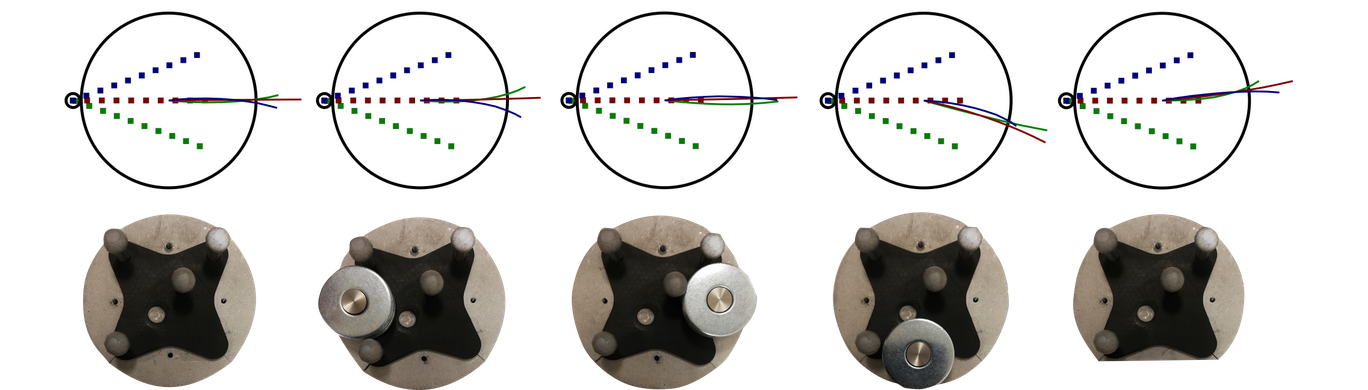}}
\label{fig:dynamics_mass}
\centering
\vspace{-3mm}
\caption{\textbf{Dynamics comparison for different mass distribution.} We compare the motion of five different objects under three different pushes (doted lines). We observe that adding a mass into the horizontal axis affects the pushes that are not straight by increasing (second case) or decreasing (third case) the tendency of the object to rotate w.r.t to the first case. Adding a mass below the horizontal axis (fourth case) displaces the center of mass downward, which makes the object rotate clock-wise for all three pushes. Instead, subtracting mass in the bottom side, as done in the last case, has the opposite effect where the center of mass is displaced upward, which makes the object rotate anti-clockwise. These results show the clear effect of mass distribution over the dynamics of pushed objects. } \label{fig:mass_dist}
\vspace{-4mm}
\end{figure*}

\textbf{Poses}: for every push we track $x,y,\theta$ of the object, and $x,y$ position of the pusher at a rate of 250~Hz. 

Making the assumption that the result of the push does not depend on the absolute pose of the object, we can remove 3 dimensions by changing to the frame of the object (which sets $x=y=\theta=0$ for the object). We can further remove another dimension by using the fact that the pusher moves at constant speed and represent the velocity (difference between final and initial pose of the pusher) with only its angle. Therefore, we can have either 3, 5 or 7 input dimensions depending on the assumptions, and the target is always 3 dimensional: $(\Delta x,\Delta y,\Delta \theta)$, the change in object pose.

\textbf{RGB-D video:} We include RGB-D recordings of the pushes at 30~Hz with  resolution 640x480. This results in a complete dataset that can be studied directly from a vision perspective or by using the accurate positions recovered by the tracking system. Given the nature of the data collection, some pushes are related because the final state of the first one corresponds to the initial state of the following push. As a result, it is possible to stitch sequences of up to 5 pushes and get longer motions of up to 5s, where the direction of the pusher changes after each second.

\subsection{Extensions to Omnipush}\label{sec:extensions}

We have included several extensions to Omnipush that can be used as extra datasets to further test generalization along different dimensions. We collected 250 pushes with the proposed protocol for the 10 objects used in \cite{yu2016more} and shown in~\figref{fig:out_objects} (which are made of steel and are much heavier). Similarly, we collected 2.5k pushes on the same surface (abs) for 5 of the 250 new objects. For a different surface made of plywood, we recorded 250 pushes for 10 of the 250 new objects and the 10 objects from~\cite{yu2016more}. These out-of-distribution datasets are intended to be used in estimating how much a given algorithm can generalize to related tasks that are outside the original distribution of tasks, checking for dataset distribution bias. 

\begin{figure}[t]
\centering
{\includegraphics[width=\linewidth]{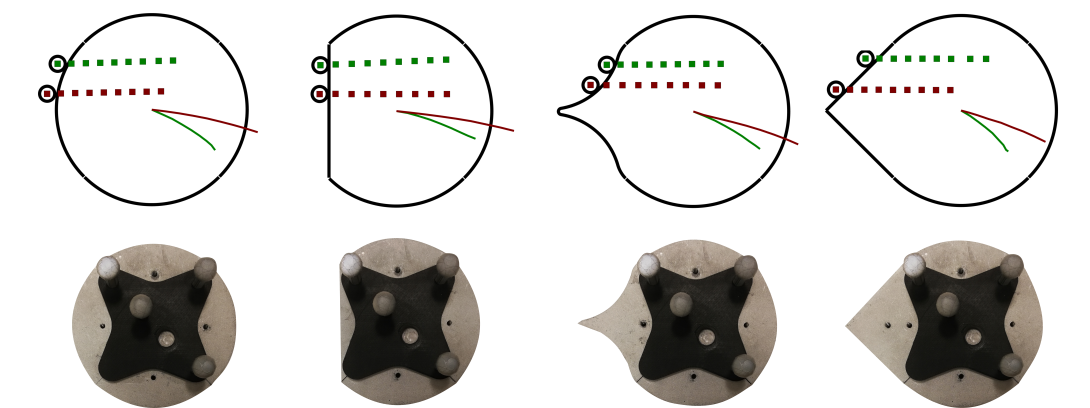}}
\centering
\vspace{-3mm}
\caption{\textbf{Dynamics comparison for different contact location.} We compare the motion of four objects that differ only on the pushed side under two different pushes (doted lines).} 
\label{fig:shape_effect}
\vspace{-5mm}
\end{figure} 

\section{Shape and weight influence over dynamics}\label{sec:influence}

There are two key factors that affect the dynamics of quasi-static pushing:
\begin{enumerate}
    \item The pressure-friction distribution, which determines the location of the center of friction.
    \item The local geometry at the contact point, which determines the Jacobian from the contact point to the center of friction (how the
contact force is transferred to the center of friction) and the
orientation of the friction cone w.r.t. the pushing direction
(the angle between the normal of the surface and the direction of the
push).  \end{enumerate} 
The Omnipush shapes were designed to explore these factors.

\subsection{Mass Distribution}

Mass distribution has a direct effect on the pressure distribution of an object and the potential to affect its dynamics. To the best of our knowledge there is little available experimental data that captures this effect. In this section, we show the effect of changing the mass distribution by adding extra weight or removing a part of the object.

The first studies of the mechanics of pushing~\citep{mason1986mechanics} already show that pressure distribution directly affects friction. This in turn affects the motion of the pushed object. \citet{Dogar2010} explored the case where all the mass of an object is at its periphery, which theoretically produces maximal rotation. With these assumptions, the authors bound possible deviations from a desired motion.

\figref{fig:mass_dist} illustrates the effect of changing pressure distribution on the trajectory of a pushed object. We compare the motion of a circle-shaped object for three different pushes (straight, up and down) and for five different mass distributions. To reduce the effect of stochasticity, we average the resulting trajectories over 10 pushes. 
As expected, the closer the added weight to the pusher, the more likely it is to rotate and deviate from a straight line. 



The results from this section show that the effect of mass distribution, which is difficult to detect just from vision, has an important effect on the dynamics of equally shaped objects. The dataset provides a useful tool to explore this effect.

\subsection{Geometry of contact edge}

The tangent to the surface at the point of pusher-object contact determines the orientation of the friction cone.
This, along with the location of the center of friction, determines the direction of the motion cone~\cite{lynch1992manipulation,hogan2016feedback}, which ultimately determines the directions along which the pusher will stick or slide on the surface of the object.
Figure~\ref{fig:shape_effect} illustrates the effect of two parallel pushes that contact differently shaped edges on an object.  Note that the flat and concave surfaces (middle two shapes) have less variation in the normal at contact and result in more similar behavior than for the circular and angled surfaces (the outer two shapes).  This effect is more subtle than the ones due to adding the weight but nevertheless important in making accurate predictions.

\section{Benchmark: meta-learning}\label{sec:modeling}


This dataset makes it possible to build predictive pushing models that generalize over object geometries and mass distributions. We present a new meta-learning benchmark for algorithms that learn to learn dynamic models for unseen objects.\looseness=-1

We consider the task of predicting the final pose of a pushed object given its initial pose and a description of the push. In particular, for this benchmark we make the assumption that dynamics are independent of the absolute pose of the object and that the magnitude of the velocity of the push is constant, which results in a 3 dimensional input: initial $x,y$ position of the pusher with respect to the object, and the angle $\phi$ of the relative velocity of the pusher to the object. From this we predict $\Delta x,\Delta y,\Delta \theta$, the change in object pose. We normalize inputs and outputs across each dimension separately to mean 0 and standard deviation 1 to make them comparable. To provide an idea of the scale of the variables we are trying to predict, we define a \textit{zero} baseline which always predict that the output will follow a Normal distribution of mean 0 and standard deviation 1, regardless of the input.\looseness=-1

We evaluate model performance using both the root mean squared error (RMSE) and the negative log-probability density error (NLPD).
To make the results more intuitive, we convert the RMSE to millimeters by multiplying it by the standard deviation of the change in position: 21.92mm (see table~\ref{table:results}). 
Note that the NLPD error measures the ability of a model at predicting the probability distribution over outcomes. For instance, our baselines predict mean and standard deviation, defining a Gaussian probability density for each input over all possible outcomes. By optimizing a NLPD loss during training, we force models to both make accurate predictions and assess their own uncertainty correctly. We compute the NLPD metric for a model $M$ and dataset $\mathcal{D}$ as: $$NLPD\left(M, \mathcal{D}\right) = -\mathbb{E}_{(x,y)\in \mathcal{D}}\left[\ln p_{M,x}(y)\right],$$
where $p_{M,x}$ is the probability distribution defined by model $M$ when given input $x$. Note that the NLPD can be negative and is lower-bounded by the differential entropy of the true (unknown) distribution.

Since pushing is experimentally stochastic, we estimate an upper-bound of the Bayes error rate for the RMSE and NLPD. This gives an approximation of the system's irreducible noise and thus a lower-bound for the models error. To compute the Bayes error, we pushed 5 random shapes 2.5k times each, and train a separate neural network (NN) per object with 2k pushes. \citet{bauza2017probabilistic} determined that when modeling the pushing dynamics, extra data helped little beyond 2k pushes for Gaussian process regression. 



To learn the dynamics, we consider two baseline algorithms: an object-independent NN that pools the data from all objects into a single dataset, and a meta-learning approach (Attentive Neural Processes~\cite{kim2018attentive}) that builds object-specific models. The details of the models, training and evaluation are in the project website. 




\myparagraph{Real-world few-shot regression.} The Omnipush dataset provides a new supervised-learning benchmark for few-shot regression. Until now, meta-learning benchmarks have mostly focused on few-shot classification~\cite{lake2015human,vinyals2016matching} or meta model-free reinforcement learning (RL)~\cite{rakelly2019efficient,humplik2019meta}, while to our knowledge few-shot regression problems have only been studied for model-based meta RL~\citep{clavera2018learning,saemundsson2018meta} and toy 1-D function datasets~\citep{finn2017model,alet2018modular}. 


By doing few-shot learning with 50 pushes for each new object, our baselines for this task achieve the results in Table~\ref{table:results}. We observe that pooling the data into a single dataset captures a sizable amount of the signal (see rows marked with ``no" meta-learning), but the meta-learning algorithm performs significantly better in terms of RMSE, halving the distance to the Bayes error rate bound for the Omnipush dataset.
This is expected in meta-learning settings where tasks share a lot of structure, but are still fundamentally different problems.

Given the results, we believe algorithms will need between 10 and 50 samples to generalize. This is because the dynamics of each push depends on the local shape of the object and thus accurate learning might require a few pushes distributed across the object's boundary.
As a consequence, we propose two benchmarks at 10 and 50 samples per new object (more details in the website), and encourage the exploration of active learning methods from meta-learned priors, to further reduce data requirements.

\myparagraph{Uncertainty estimation.} To the best of our knowledge, Omnipush is the first standarized benchmark for uncertainty estimation in meta-learning. This is important because having few data points about a new task leads to an intrinsic epistemic uncertainty, since we cannot be sure about our model from a small amount of data.
While there has been a growing interest in meta-learning algorithms that provide uncertainty estimates~\cite{finn2018probabilistic,edwards2016towards,garnelo2018conditional,gordon2018meta, yoon2018bayesian}, there was no standard benchmark to measure progress on that front. 
Moreover, uncertainty in Omnipush is particularly interesting because there is a non-negligible amount of irreducible noise, which is also known to be heteroscedastic~\cite{bauza2017probabilistic}, i.e., some pushes are noisier than others. \looseness=-1

From our results, we see that there is still a lot of progress to be made with respect to uncertainty estimation in meta-learning. Despite good RMSE scores, the meta-learning baseline has NLPD scores that are similar to those for the non meta-learning baseline and much worse than those of the Bayes upper-bound. This suggests that the current method is unable to accurately asses its own uncertainty. 

\myparagraph{Generalization beyond meta-training distribution.} This dataset aims at providing a tool to learn general models. To show that, we test the previous algorithms on the out-of-distribution datasets from section~\ref{sec:extensions}.
For the three datasets, the meta-learning baseline shows some (limited) capacity to adapt in terms of RMSE and NLPD. Moreover, there is a significant gap between the performance on out and in distribution objects, since meta-training datasets come from the latter distribution. New algorithms need to be designed for better generalization outside of the meta-training task distribution, specially in the context of uncertainty estimation.

\begin{table}[t]
  {\centering
  \vspace{8pt}
  \tabcolsep=0.1cm{
  \caption{}\label{table:results}
  \begin{tabular}{llrrr}
    \toprule[1pt]
      \textbf{Dataset}&Meta-learning& NLPD & RMSE  & Dist. equivalent\\ 
    \midrule[1pt]
      Zero: $Normal(0,1)$ & -- & 4.25 & .997 & 21.9 mm\\
      \midrule[0.3pt]
      Bound on Bayes error& -- & $\scriptstyle <$-2.15& $\scriptstyle <$.165 & 3.6 mm\\
      \midrule[0.3pt]
      \multirow{ 2}{*}{\textbf{Omnipush}} & no & 0.16& .328 & 7.2 mm\\
      & yes \cite{kim2018attentive} & -0.11& .225 & 4.9 mm \\
      \midrule[0.3pt]
      \multirow{ 2}{*}{\textbf{Out-of-distribution}}& no& 2.46 & .512 & 11.2 mm\\ 
      &yes \cite{kim2018attentive} &  2.33 & .469 & 10.3 mm\\
      \midrule[0.3pt]
      \multirow{ 2}{*}{Different surface }& no& 1.85 & .333 & 7.3 mm\\ 
      &yes \cite{kim2018attentive} & 1.16& .285 & 6.2 mm\\
      \midrule[0.3pt]
      \multirow{ 2}{*}{Different objects }&no& 2.80& .601 & 13.2 mm\\
      &yes \cite{kim2018attentive} & 3.09& .558 & 12.2 mm\\
      \midrule[0.3pt]
      \multirow{ 2}{*}{Diff. obj. \& diff. surf.}&no & 2.72& .562 & 12.3 mm\\
      &yes \cite{kim2018attentive}& 2.73& .517 & 11.3 mm\\
    \bottomrule[1pt]
  \end{tabular}}
  \newline
  \newline}
  Horizontal lines separate different datasets, and for simplicity, \textit{out-of-distribution} agglomerates the three out-of-distribution datasets into a single benchmark. We will keep up-to-date tables at \url{https://web.mit.edu/mcube/omnipush-dataset/} and welcome submissions. 
  \vspace{-0.5\baselineskip}
\end{table}

\section{Benchmark: video prediction}\label{sec:prediction}

To characterize the stochastic nature of the planar pushing system, we evaluate stochastic video prediction methods based on variational autoencoders (VAEs) in both action-free and action-conditional settings. In the action-free setting, the goal is to predict future frames $x_{c+1:T}$ conditioned on the initial frames $x_{1:c}$, where $c$ is the number of initial frames and $T$ is the horizon of the push. In the action-conditional setting, the model is additionally conditioned on robot arm's action sequences $a_{1:T-1}$ throughout the push. Code and pretrained models will be released on the project's website.

\begin{figure*}[t]
\centering
{\includegraphics[width=\linewidth]{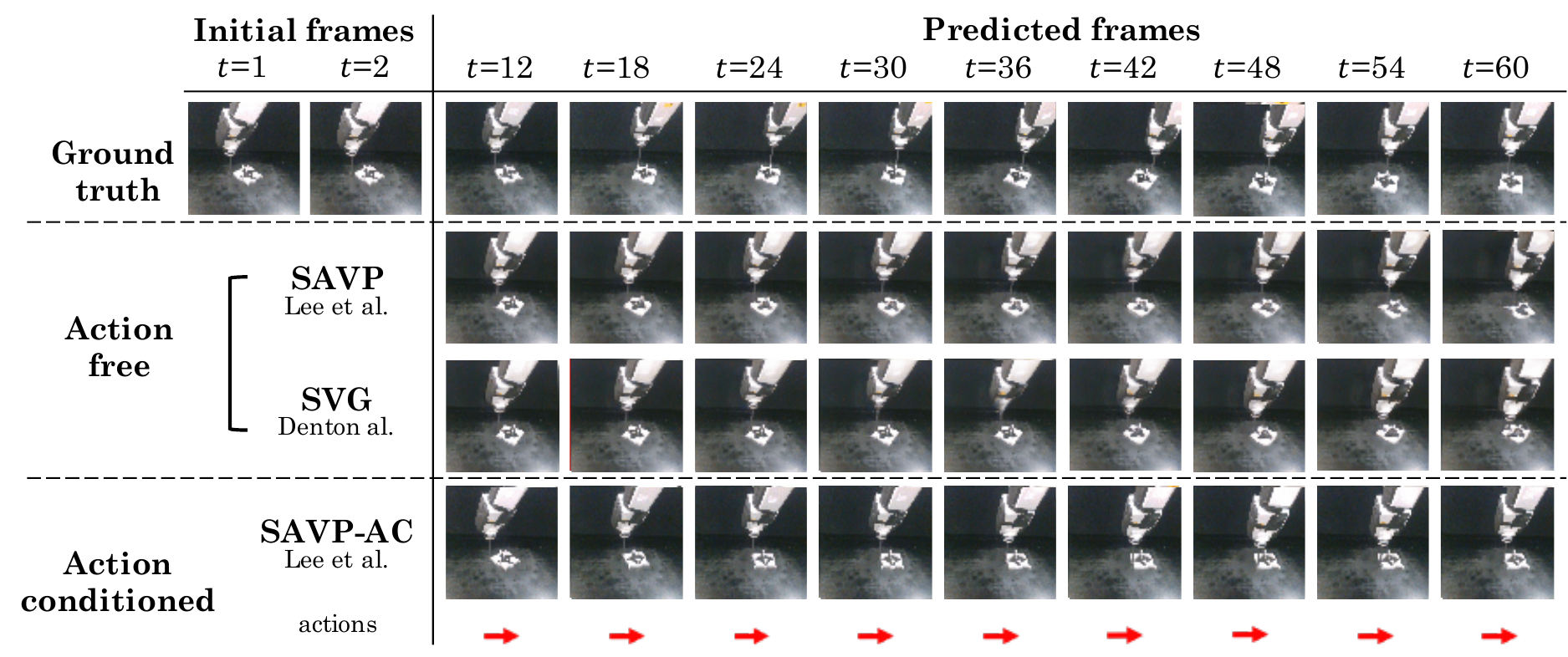}}
\centering
\vspace{-7mm}
\caption{\textbf{Qualitative results for video prediction.} (top) Ground truth video sequence. (middle) We show qualitative comparisons between baselines in the action-free setting. All models are conditioned on the first two frames of unseen test videos. SVG preserves object's shape better, while models trained with a GAN loss generate more accurate motion. (bottom) We show qualitative results in the action-conditional setting. The red arrows represent the robot arm's pushing direction at each time step. The generated video simulates the future according to randomly sampled actions and is therefore different from the ground truth video although they share the same first two frames. This demonstrates that the model is sensitive to the actions it is conditioned on, since it can simulate counterfactual action sequences (what would have happened if the actions had been other than they actually were).}
\label{fig:video_pred_qual}
\end{figure*} 

We compare the following methods on our dataset in the action-free setting. 

\begin{itemize}
    \item \textbf{SVG-LP.~\cite{denton2018}} An improved VAE model with a learned prior that can dedicate its capacity to model stochastic dynamics. The generator consists of an LSTM~\cite{hochreiter1997long} and separate convolutional networks as encoder-decoder.
    \item \textbf{SAVP.~\cite{lee2018}} A VAE-GANs model that is trained with both variational lower bound and adversarial loss. Different from SVP-LP, the generator is a convolutional LSTM~\cite{xingjian2015convolutional}.
\end{itemize}

\begin{figure}
\centering
{\includegraphics[width=\linewidth]{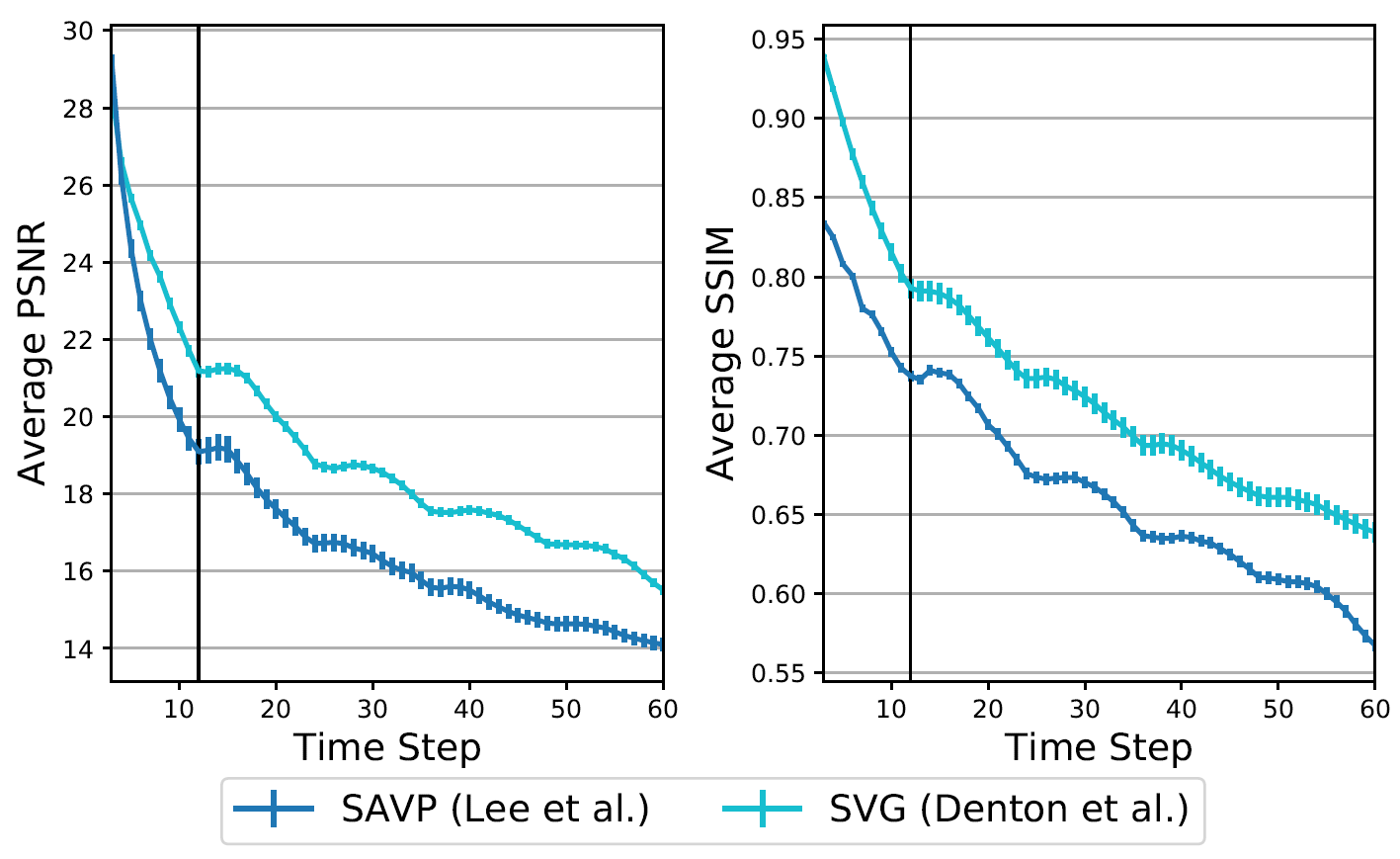}}
\centering
\vspace{-7mm}
\caption{\textbf{Quantitative results for video prediction in the action-free setting.} We show quantitative results of baselines in the action-free setting. All models are conditioned on the first two frames of unseen test videos. SVG performs slightly better than SAVP. However, previous work~\cite{lee2018, zhang2018unreasonable} has pointed out that PSNR and SSIM correspond poorly to human judgments, and GANs, despite achieving perceptually realistic results, are expected to perform poorly on these distortion-based metrics~\cite{blau2018perception}. Mean SSIM and PSNR over test videos is plotted with $95\%$ confidence interval. Models were trained to predict up to 12 frames into the future (black line) but at test time can be used to predict even farther.}
\label{fig:video_pred_quan}
\vspace{-6mm}
\end{figure}

We follow the data preprocessing steps from previous work~\citep{lee2018}. First, we center-crop each frame to a 480x480 square and resize it to a spatial resolution of 64x64. In our experiment, we condition on the first 2 frames and train the model to predict 12 frames in the future, which corresponds to length of the push, 1 second. All models are trained with Adam optimizer~\cite{Kingma2014AdamAM} for 300k iterations with learning
rate of $0.001$ and batch size of 32. 

We evaluate the models by sampling 100 videos, then calculate the Peak Signal-to-Noise Ratio (PSNR) and structural similarity (SSIM)~\cite{wang2004image} scores between the best generated video from those samples and the ground truth as in~\citep{denton2018,lee2018}. The  qualitative and quantitative results are shown in~\figref{fig:video_pred_qual} and~\figref{fig:video_pred_quan}. We observe that state-of-the-art video prediction models don't capture the object's shape as well in later frames as in earlier frames, and it tends to become circular in later predicted frames. This motivates future research in video prediction models that can condition on additional object information.

\section{Conclusion}
\label{sec:conclusion}
This paper presents Omnipush, a high-fidelity experimental dataset of planar pushing interactions with multiple sensory inputs and large variety of objects. In total, the dataset contains 250 pushes for 250 different objects, ground truth trajectories and RGB-D video of both the robot and the pushed object. As a result, this dataset opens the possibility to study challenging dynamics in the context of frictional manipulation. 

This dataset can be used to study multiple technical problems. In Sections~\ref{sec:modeling} and~\ref{sec:prediction} we detailed two: meta-learning dynamic models and video prediction, and highlighted some of the limitations of state-of-the-art algorithms, such as bad performance at quantifying uncertainty and generalization to out-of-distribution tasks. We envision more uses and problems that can be addressed with this dataset. For instance:

\myparagraph{Finding the ``right" representation.} An important open problem in robotics is to find task-aware representations. Ideally, these should be structured enough to facilitate generalization and planning, but also flexible enough to accommodate a wide variety of objects and sensory streams. Omnipush provides the two extremes of such representations (poses and pixels) and a structured variety of objects that eases interpretation and comparison of different approaches.

\myparagraph{Meta-learning, uncertainty and generalization.} Omnipush provides a different benchmark for the meta-learning community, which is currently mostly dominated by image classification benchmarks. In particular, its low dimensionality enables faster experimentation and easier interpretations of the models. Moreover, the inclusion of uncertainty and out-of-distribution datasets provides new challenges for the community.

\myparagraph{Leveraging object models for learning.} Omnipush objects come from a well defined distribution, which facilitates works that use a model of the object. For instance, in~\cite{alet2019graph} we use Ominpush to create a dynamic model out of an object from the top-down view of its CAD model.


\myparagraph{Filtering.} Real-time accurate filtering is key in real robotic scenarios with noisy dynamics and observations. Because this dataset provides multiple sensory streams to describe the object and pusher poses, it is possible to test filtering algorithms that use raw images as the sensory information and compare them with ground truth from tracking. 

\myparagraph{Action-conditional video-prediction.} In the intersection of robotics and computer vision, we hope Omnipush sparks research in more structured predictive models that go beyond predicting textured movements in pixel space. In particular we are interested in better models for action-conditional prediction that work reliably even for long horizons.

In future work, we want to explore some of the mentioned challenges and provide extensions to the dataset. Among the extensions, it would be interesting to cover interactions between a robot and multiple objects, objects of different materials, 3d interactions, and soft or articulated objects.  
\bibliographystyle{IEEEtranN} 
{\footnotesize \bibliography{references}} 
\end{document}